\documentclass[12pt, letterpaper]{article}
\usepackage[margin=1in]{geometry}

\usepackage[square,,numbers]{natbib}

\usepackage{graphicx,amssymb,amsmath}
\usepackage{array}
\usepackage{balance}
\usepackage{subcaption}
\usepackage{nicefrac}

\def\bbR{{\mathbb R}}

\renewcommand{\cite}[1]{\citep{#1}}
\newcommand*{\affaddr}[1]{#1} 
\newcommand*{\affmark}[1][*]{\textsuperscript{#1}}

\title{The Effectiveness of Instance Normalization: \\a Strong Baseline for Single Image Dehazing}

\author{
Zheng Xu\affmark[1]\thanks{xuzh@cs.umd.edu}, \quad Xitong Yang\affmark[1], \quad Xue Li\affmark[2], \quad  Xiaoshuai Sun\affmark[3]\\
\affaddr{\affmark[1]University of Maryland, College Park}\\
\affaddr{\affmark[2]Facebook, Menlo Park}\\
\affmark[3] Harbin Institute of Technology, China \\
}

\usepackage{cleveref}

\begin{document}

\sloppy

\maketitle

\begin{abstract}
We propose a novel deep neural network architecture for the challenging problem of single image dehazing, which aims to recover the clear image from a degraded hazy image. 
Instead of relying on hand-crafted image priors or explicitly estimating the components of the widely used atmospheric scattering model, our end-to-end system directly generates the clear image from an input hazy image. 
The proposed network has an encoder-decoder architecture with skip connections and instance normalization. 
We adopt the convolutional layers of the pre-trained VGG network as encoder to exploit the representation power of deep features, and demonstrate the effectiveness of instance normalization for image dehazing. 
Our simple yet effective network outperforms the state-of-the-art methods by a large margin on the benchmark datasets. 
\end{abstract}


\section{Introduction}

Images captured in the wild are often degraded in visibility, colors, and contrasts caused by haze, fog and smoke. 
Recovering high-quality clear images  from degraded images (a.k.a. image dehazing) is beneficial for both low-level image processing and high-level computer vision tasks. 
Dehazed images are more visually appealing to generate for image processing tasks.
Dehazed images can improve the robustness of vision systems that often assume clear images as input. 
Typical applications that benefit from image dehazing include image super-resolution, visual surveillance, and autonomous driving. 
Image dehazing is highly desired because of the increasing demand of deploying visual system for real-world applications. 

Image dehazing is a challenging problem. The effect of haze is caused by atmospheric absorption and scattering that depend on the distance of the scene points from the camera. 
In computer vision, the hazy image is often described by a simplified physical model, i.e., the atmospheric scattering model \cite{mccartney1976optics,narasimhan2002vision,he2011single,li2017reside},
\begin{equation}
I(x) = J(x)t(x) + A(1-t(x)), \label{eq:phy}
\end{equation}
where $I(x)$ is the observed hazy image, $J(x)$ is the scene
radiance (clear image), $t(x)$ is the medium transmission
map, and $A$ is the global atmospheric light. 
When the atmosphere is homogeneous, $t(x)$ can be further expressed
as a function of the scene depth d(x) and the scattering
coefficient $\beta$ of the atmosphere as
$t(x) = \exp(-\beta d(x))$. 
The goal of image dehazing is to recover clear image $J(x)$ from hazy image $I(x)$. 
Single image dehazing is particularly challenging. It is under-constrained because haze is dependent on many factors,  including the unknown depth information that is difficult to recover from a single image.



The atmospheric scattering model \eqref{eq:phy} has been extensively used in previous methods for single image dehazing
\cite{fattal2008single,tan2008visibility,tarel2009fast,he2011single,meng2013efficient,fattal2014dehazing,berman2016non,chen2016robust}.
These works either separately or jointly estimate the  transmission map $t(x)$ and the  atmospheric light $A$ to generate the clear image from a hazy image. 
 Due to the under-constrained nature of single image dehazing, the success of previous methods often relies on hand-crafted priors such as  dark channel prior \cite{he2011single},  contrast
color-lines \cite{fattal2014dehazing}, color attenuation prior \cite{zhu2015fast}, and non-local pior \cite{berman2016non}. 
 However, it is difficult for these priors to be always satisfied in practice. For example,  dark channel prior is known to be unreliable for areas that are similar to the atmospheric light. 

More recent works learn convolutional neural networks (CNNs) to estimate components in the atmospheric scattering model for image dehazing \cite{cai2016dehazenet,ren2016single,li2017aod,li2018cascaded,zhang2018densely,yang2018towards}. These methods are often trained with limited (synthetic) images, and use only a few layers of convolutional filters. The learned shallow networks have limited capacity to represent or process images, making them difficult to surpass the prior-based methods.
In contrast, training deep neural networks with large-scale data has made significant progress and achieved state-of-the-art performance in many vision tasks \cite{krizhevsky2012imagenet,simonyan2014very,he2016deep}. Moreover, the deep features extracted by a pre-trained deep network are used as powerful image representation in many applications, such as domain invariant recognition \cite{donahue2014decaf}, perceptual evaluation \cite{zhang2018unreasonable}, and characterizing image statistics \cite{gatys2016image}. More recently, the architecture of CNNs itself has been recognized as a prior for image processing \cite{ulyanov2017deep}. In this paper, we study how to release the power of \emph{deep} network  for single image dehazing. 




We propose an encoder-decoder architecture as an end-to-end system for single image dehazing. 
We exploit the representation power of deep features by adopting the convolutional layers of the deep VGG net \cite{simonyan2014very} as our encoder, and pre-train the encoder on large-scale image classification task \cite{russakovsky2015imagenet}. 
We add skip connections with instance normalization between the encoder and decoder, and then train decoder with both $\ell_2$ reconstruction loss and VGG perceptual loss \cite{zhang2018unreasonable}. 
We show that the recently proposed instance normalization \cite{ulyanov2016instance}, which is designed for image style transfer, is also effective in image dehazing.   
The proposed method effectively learns the statistics of clear images based on the deep feature representation, which benefits the dehazing process on the input image.
Our approach outperforms the state-of-the-art results by a large margin on a recently released benchmark dataset \cite{li2017reside}, and performs surprisingly well in several cross-domain experiments. 
Our method depends on neither the explicit atmospheric scattering model nor the hand-crafted image priors, and only exploits the deep network architecture and pre-trained models to tackle the under-constrained single image dehazing problem. 
Our simple yet effective network can serve as a strong baseline for future study in this topic.



\section{Related work}
\label{sec:related}
Traditional methods focus on representing human knowledge as priors for image processing. 
\citet{tan2008visibility} assumes higher contrast of clear images and
proposes a patch-based contrast-maximization method. 
\citet{fattal2008single} assumes the transmission and surface shading are locally uncorrelated, and estimates the albedo of the scene. 
Dark channel prior (DCP) ~\cite{he2011single} assumes local patches contain low intensity pixels in at lease one color channel and hence estimates the transmission map.
Fast visibility restoration (FVR) ~\cite{tarel2009fast} is a filtering approach by atmospheric veil inference and corner preserving smoothing. 
\citet{meng2013efficient} uses boundary constraint and contextual regularization (BCCR), and 
\citet{chen2016robust} uses gradient residual minimization (GRM) to surpress artifacts.  \citet{tang2014investigating} combines priors by learning with random forests model. Color attenuation prior (CAP) \cite{zhu2015fast} assumes a linear model of brightness and the saturation and then learns the coefficients.  \citet{berman2016non} assumes each color cluster in the clear image becomes a line in RGB space, and  proposes non-local image dehazing (NLD). 

There is an increasing interest in applying convolutional neural networks (CNNs) for image dehazing. DehazeNet~\cite{cai2016dehazenet} and multi-scale convolutional neural networks (MSCNN) \cite{ren2016single} are trained to estimate the transmission map.  
AOD-Net\cite{li2017aod} estimates a new variable based on the transformation of the atmospheric scattering model. 
\citet{zhang2018densely} and \citet{li2018cascaded} estimate transmission map and atmospheric light by separate CNNs. \citet{yang2018towards} adversarially train generators for components of  the atmospheric scattering model. 
These methods use relatively small CNNs and do not exploit the pre-trained deep networks for image representation.  
A few days before our submission, we notice a preprint \cite{cheng2018semantic} that also uses the pre-trained deep networks. 
The proposed method is quite different from \cite{cheng2018semantic}: we use encoder-decoder with skip connections, while \citet{cheng2018semantic} only use feature maps extracted from one layer of the pre-trained network as input; we study instance normalization and demonstrate its effectiveness; we train an end-to-end system from hazy image to clear image, while \citet{cheng2018semantic} estimate  transmission map and atmospheric light; we can generate impressive results without explicitly applying the atmospheric scattering model. 

Deep neural networks can be used as ``priors'' for image generation and image processing. The architecture of CNNs itself can be a constraint for image processing \cite{ulyanov2017deep} and image generation \cite{kingma2013auto,goodfellow2014generative}. A pre-trained deep networks can be used as general purpose feature extractors \cite{donahue2014decaf} and perceptual metric \cite{zhang2018unreasonable}. The second-order information of the features extracted by a pre-trained network describes the style of  images \cite{gatys2016image}. Instance normalization layers that effectively change the statistics of deep features are widely used for image style transfer \cite{ulyanov2016instance,dumoulin2016learned,ghiasi2017exploring,huang2017arbitrary}. 






\begin{figure}[t]
\centerline{
\includegraphics[width=0.85\linewidth]{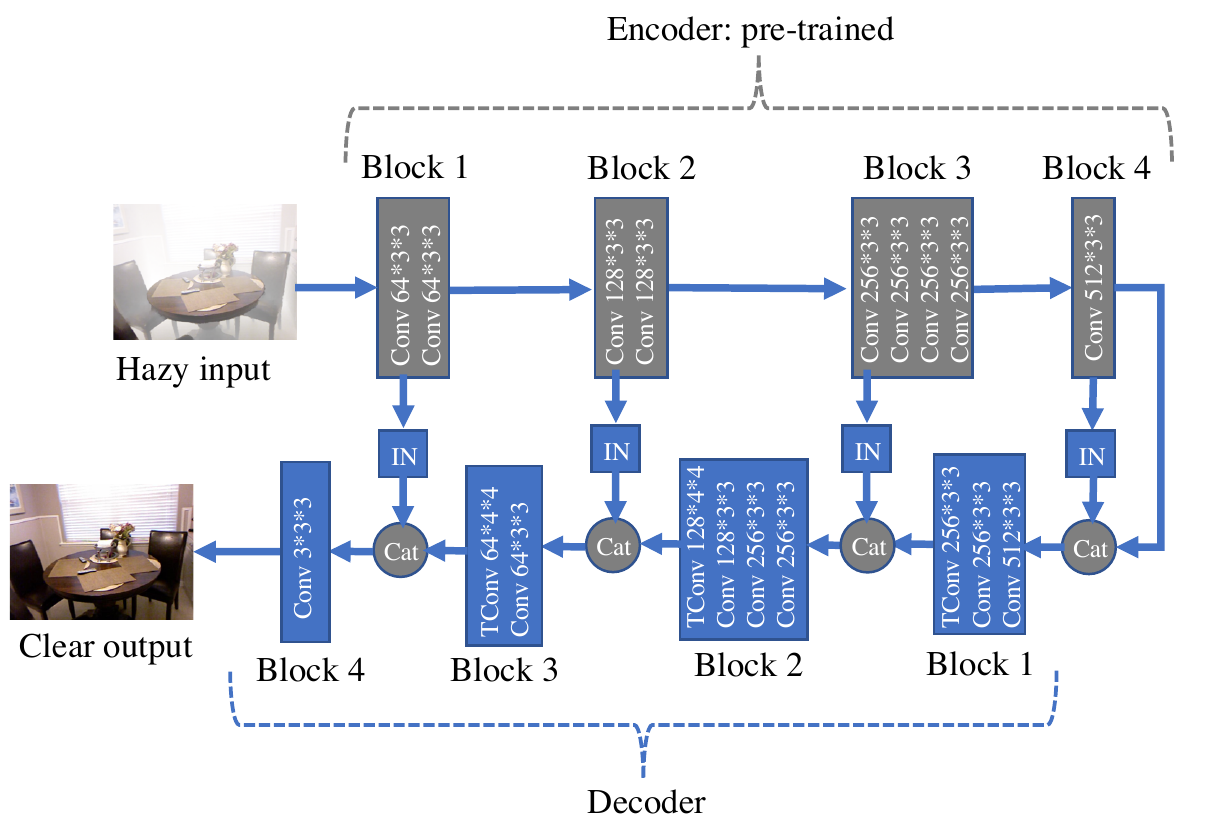}
}
\vspace{0.2cm}
\caption{
The proposed network: encoder-decoder with skip connections and instance normalization (IN); convolutional layers of pre-trained VGG~\cite{simonyan2014very}  are used as encoder; $\ell_2$ reconstruction loss and VGG perceptual loss are used for training decoder and IN layers.}
\label{fig:net}
\end{figure}

\section{VGG-based U-Net with instance normalization}


We propose an end-to-end encoder-decoder network architecture for single image dehazing, as shown in \cref{fig:net}. The input is a hazy image, and the output is the desired clear image. We introduce different components of the network in the following paragraphs of this section. 

\textbf{Encoder. }  
Our encoder uses the convolutional layers of the VGG net \cite{simonyan2014very} pre-trained on Imagenet large-scale image classification task \cite{russakovsky2015imagenet}. VGG net contains five blocks of convolutional layers, and we use the first three blocks and the first convolutional layer of the forth block. Each block contains several convolutional layers, and each convolutional layer is equipped with ReLU \cite{krizhevsky2012imagenet} as activation function. The width (number of channels) and size (height and width) of convolutional layers are shown in \cref{fig:net}. There is a maxpooling layer of stride two between blocks, which enlarges the receptive field of higher layers. The width of convolutional layer is doubled after the subsampling of feature maps by maxpooling.

 The pre-trained VGG net is a powerful feature extractor for perceptual metric \cite{zhang2018unreasonable} and image statistics \cite{gatys2016image}. 
Our encoder is deep and wide, and the extracted deep features are capable to capture the semantic information of the input image.  We fix the encoder during training to exploit the power of pre-trained VGG net as ``priors'', and avoid overfitting from relatively small number of samples in image dehazing dataset.  

\textbf{Decoder and skip connection. } 
Our decoder is designed to be roughly symmetric to the encoder. The decoder also contains four blocks, and each block contains several convolutional layers. The last layer of the first three blocks of the decoder uses transposed convolutional layer to upsample the feature maps. We use ReLU activation for convolutional and transposed convolutional layers except for the last layer, where we use Tanh as activation function. 

We add skip connections from the output of the first convolutional layer of encoder block 1,2,3 to the input of decoder block 4,3,2 by concatenating (cat) the feature maps, respectively.  
Hence our deep encoder-decoder network has a U-Net \cite{ronneberger2015u,isola2016image} structure except that our skip connections are based on blocks instead of layers . We use trainable instance normalization for skip connections, and have instance normalization before each convolutional layer in decoder except the first one. 
Our deep encoder-decoder network has large capacity, and skip connections make the information smoothly flow to easily train a large network. 

\textbf{Instance normalization. }
We briefly review instance normalization \cite{ulyanov2016instance}, and discuss our motivation in applying instance normalization for single image dehazing. 
Let $x\in \bbR^{N\times C\times H\times W}$ represent the feature map of a convolutional layer from a minibatch of samples, where $N$ is the batch size, $C$ is the width of the layer (number of channels), $H$ and $W$ are height and width of the feature map. $x_{nchw}$ denotes the element at height $h$, width $w$ of the $c$th channel from the $n$th sample, and instance normalization layer can be written as,
\begin{equation}
\begin{split}
IN(x_{nchw}) & = \gamma_{nc} \left( \frac{x_{nchw} - \mu_{nc}}{\sigma_{nc}} \right) + \beta_{nc}, \text{ where }   \\
 \mu_{nc} & = \frac{1}{HW} \sum_{h=1}^{H} \sum_{w=1}^{W} x_{nchw}, \ \sigma_{nc} = \sqrt{\frac{1}{HW} \sum_{h=1}^{H} \sum_{w=1}^{W} (x_{nchw}-\mu_{nc})^2 + \epsilon},
\end{split}
\end{equation}
$\gamma_{nc}, \beta_{nc}$ are learnable affine parameters, $\epsilon$ is a very small constant, and $\mu_{nc}, \sigma_{nc}^2$ represent the mean and variance for each feature map per channel per sample.

If we replace instance level variables $\gamma_{nc}, \beta_{nc}, \mu_{nc}, \sigma_{nc}^2$ with batch level variables $\gamma_{c}, \beta_{c}, \mu_{c}, \sigma_{c}^2$ that are estimated for all samples of a minibatch, we get the well-known batch normalization layer \cite{ioffe2015batch}. We show instance normalization is preferred than batch normalization for single image dehazing in our experimental ablation study.

The learnable affine parameters $\gamma_{nc}, \beta_{nc}$ of instance normalization shift the first and second order statistics (mean and variance) of the feature maps. Instance normalization is effective for image style transfer, and the style of images can be represented by learned affine parameters \cite{dumoulin2016learned}. Shifting the statistics of deep features extracted by pre-trained networks has achieved impressive results for arbitrary style transfer \cite{huang2017arbitrary}. Shifting the statistics of images is intuitive for dehazing, however, it can be difficult to decide the exact amount to change because haze depends on the unknown depth. The deep features extracted by a pre-trained VGG net contain semantic information to effectively infer depth for haze, and hence the learned affine parameters effectively shift the statistics of images. We apply instance normalization on the deep features extracted by pre-trained VGG net for single image dehazing.

\textbf{Training loss. }
Our network is trained with both reconstruction loss and VGG perceptual loss. Denoting the training pairs of hazy image and clear image as $(I_n, T_n), n=1,\ldots, N$, we use the mean squared loss,
\begin{equation}
\min_{F} \frac{1}{N} \sum_{n=1}^{N} \| F(I_n) - T_n \|^2 + \lambda \| g(F(I_n)) - g(T_n) \|^2,  \label{eq:loss}
\end{equation} 
where $F$ represents the trainable instance normalization and decoder layers in our network, $g$ represents the perceptual function, and $\lambda$ is a hyperparameter. We set $\lambda=1$ , and use the features extracted by the first convultional layer of the third block from the pre-trained VGG net as perceptual function.

\section{Experiments}
In this section, we conduct various experiments on both synthetic and natural images to demonstrate the effectiveness of the proposed method. The atmospheric scattering model is widely used to synthesize images for both training and testing. The hazy images are synthesized from groundtruth clear images and grountruth depth images \cite{li2017reside,ancuti2016d}, or estimated depth images \cite{sakaridis2017semantic}. 

We train our model on the recently released RESIDE-standard dataset  \cite{li2017reside}. RESIDE-standard contains 13,990 images for training, and 500 images for testing. These images are generated by existing indoor depth datasets, NYU2~\cite{silberman2012indoor} and Middlebury stereo~\cite{scharstein2014high}.  The atmospheric scattering model is used, where atmospheric lights $A$ is randomly chosen between (0.7, 1.0) for each channel, and scattering
coefficient $\beta$ is randomly selected  between (0.6, 1.8). 

We also apply our model trained on RESIDE-standard for cross-domain evaluation on D-Hazy~\cite{ancuti2016d}, I-Haze~\cite{ancuti2018ihaze}  and O-Haze~\cite{ancuti2018ohaze} dataset. D-Hazy dataset \cite{ancuti2016d} is another synthetic dataset, which contains  23 images synthesized from Middlebury and 1449 images synthesized from NYU2, with atmospheric lights $A=(1,1,1)$ and scattering coefficient $\beta=1$. Though D-Hazy dataset use the same clean images as RESIDE-standard, the generated hazy images are quite different. I-Haze~\cite{ancuti2018ihaze}  and O-Haze~\cite{ancuti2018ohaze} are two recent released datasets on natural indoor and outdoor images, respectively. I-Haze contains 35 pairs of indoor images and O-Haze contains 45 pairs of outdoor images, where the hazy images are generated by using a physical haze machine. 

We compare our results quantitatively and qualitatively with previous methods.  We compare with prior-based methods, DCP~\cite{he2011single}, FVR~\cite{tarel2009fast}, BCCR~\cite{meng2013efficient} , GRM~\cite{chen2016robust}, CAP~\cite{zhu2015fast} and NLD~\cite{berman2016non} . We also compare with learning-based methods DehazeNet~\cite{cai2016dehazenet}, MSCNN~\cite{ren2016single} ,  and AOD-Net~\cite{li2017aod}. We have provided a brief review of these baseline methods in \cref{sec:related}. We use peak signal-to-noise ratio (PSNR) and structural similarity (SSIM)  as metrics for quantitative evaluation. For the benchmark evaluation on RESIDE-side, all the learning-based methods are trained on the same dataset. For cross-domain evaluation on D-Hazy, O-Haze and I-Haze, we use the released best pre-trained model for the learning-based baseline methods. 

We train our model by SGD with minibatch size 16 and learning rate 0.1 for 60 epochs, and linearly decrease the learning rate after 30 epochs. We use momentum 0.9 and weight decay $10^{-4}$ for all our experiments. We will release our Pytorch code and pre-trained models.  

\subsection{Quantitative evaluation on benchmark dataset}

\begin{table}[t]
\centering
\caption{Quantitative results on RESIDE-standard dataset \cite{li2017reside}.} \vspace{0.3cm}
\begin{tabular}{|c|c|c|c|c|c|c|c|c|c|}
\hline
 & DCP~\cite{he2011single} & FVR~\cite{tarel2009fast} & BCCR~\cite{meng2013efficient} & GRM~\cite{chen2016robust} & CAP~\cite{zhu2015fast} \\ 
\hline
PSNR & 16.62 & 15.72 & 16.88 & 18.86 &19.05   \\
SSIM & 0.8179 & 0.7483 & 0.7913 & \underline{0.8553} & 0.8364 \\
\hline
& NLD~\cite{berman2016non} & DehazeNet~\cite{cai2016dehazenet} & MSCNN~\cite{ren2016single} & AOD-Net~\cite{li2017aod} & \textbf{Ours} \\
\hline
PSNR & 17.29 & \underline{21.14} & 17.57 & 19.06 &  \textbf{27.79}\\
SSIM & 0.7489 & 0.8472 & 0.8102 & 0.8504 & \textbf{0.9556}\\
\hline
\end{tabular}%
\label{tab:soa}%
\vspace{-0.3cm}
\end{table}%

We present the performance of our network and baseline methods on the RESIDE-standard benchmark dataset \cite{li2017reside} in \cref{tab:soa}. Our network and the learning-based baselines \cite{cai2016dehazenet,ren2016single,li2017aod} are trained on the provided synthetic data, and evaluated on the separate testing set.  We evaluate our results by metrics provided by \cite{li2017reside}, and compare with the baseline results  reported in \cite{li2017reside}. The learning-based methods perform slightly better than the prior-based method. CAP~\cite{zhu2015fast} performs best in prior-based method, which has a learning phase for the coefficients of the linear model. DehazeNet~\cite{zhu2015fast} performs best in baseline methods, which uses a relatively small network to predict components. 

Our approach outperforms all the baseline methods on both PSNR and SSIM by a large margin.  The synthetic data for both training and testing are generated by the atmospheric scattering model, and the baseline methods explicitly use the atmospheric scattering model. In contrast, our approach only uses instance normalization to transform the statistics of deep features . The superior performance of our network on the benchmark dataset demonstrate the effectiveness of \emph{deep} networks and instance normalization for single image dehazing. 

\subsection{Ablation study}

\begin{table}[tbhp]
\centering
\caption{Ablation study on RESIDE-standard dataset. } \vspace{0.3cm}
\begin{tabular}{|c|c|c|c|c|c|c|c|c|c|}
\hline
Skip & NA & BN & IN & NA & BN \\ 
\hline
Dec & NA & NA & NA & BN & BN \\ 
\hline
PSNR & 18.24 &  25.67 &  26.00 & 25.99  &26.38  \\
SSIM & 0.7945 &  0.9442 & 0.9414   &  0.9385 & 0.9519 \\
\hline
\hline
 Skip & IN  & NA & BN & IN & Perceptual\\ 
 \cline{1-5}
 Dec & BN & IN & IN & IN  & loss \\ 
\hline
PSNR &  26.89 & 26.57 &  27.67& \underline{27.75} &  \textbf{27.79}\\
SSIM &  0.9535  & 0.9381 &  0.9543& \underline{0.9549} & \textbf{0.9556}\\
\hline
\end{tabular}%
\label{tab:ablation}%
\vspace{-0.2cm}
\end{table}%

We provide more discussion on the proposed network. We verify the effectiveness of instance normalization with ablation study on network structures, as shown in \cref{tab:ablation}. 
We use no normalization (NA), batch normalization (BN), or instance normalization (IN) for skip connections and decoders, respectively. 
The normalization layers are added before each convolutional layer of the decoder except for the first layer. 
All the results in \cref{tab:ablation} are obtained by only using reconstruction loss ($\lambda=0$ in loss function \eqref{eq:loss}) except for the last one, where  IN and combined loss ($\lambda=1$) are used. We train and evaluate our network on the RESIDE-standard dataset. 

First, comparing the NA results in \cref{tab:ablation} with previous best results in \cref{tab:soa}, our encoder-decoder only achieves competitive results. 
Second, adding normalization to either skip connections or decoder  significantly improves the performance of our network. 
The normalization layers for decoder are implicitly applied to the features from the skip connections, which makes the result of only normalizing decoder slightly better than only normalizing skip connections.
 Third, instance normalization works better than batch normalization, which demonstrates the effectiveness of shifting the mean and variance of deep features at instance level. 
 
\begin{figure}[t]
\vspace{-0.1cm}
\centerline{
\includegraphics[width=0.95\linewidth]{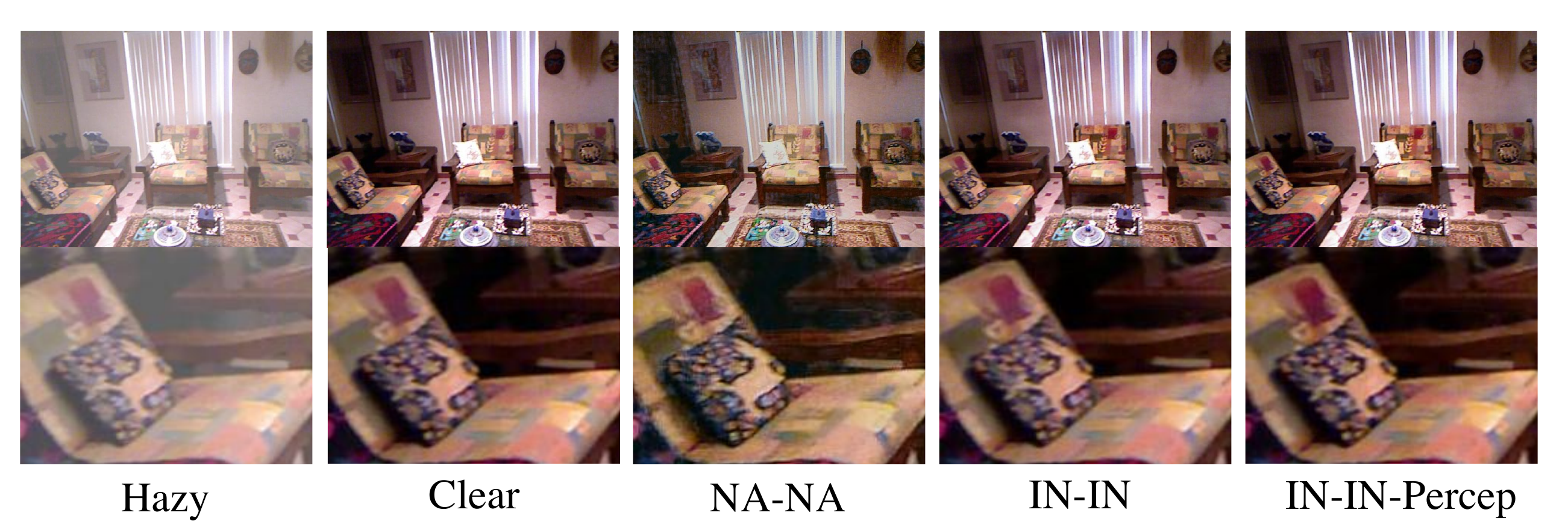}
}
\vspace{0.2cm}
\caption{
An example of qualitative results in ablation study. We zoom in the bottom left corner of the images to show more details in the second row.}
\label{fig:abla}
\vspace{-0.3cm}
\end{figure}

 Finally, the perceptual loss only helps a little for quantitative evaluation, but it can help generate more visually appealing output images. We show an qualitative example in \cref{fig:abla}, where the hazy input, the groundtruth clear image, outputs of our network without normalization layers and no perceptual loss (NA-NA), our network with instance normalization and no perceptual loss (IN-IN), and our network with instance normalization and perceptual loss (IN-IN-Percep). We enlarge the bottom left corner of the results to show more details. The results of IN-IN look much better than NA-NA. The enlarged area of the result with perceptual loss (IN-IN-Percep) looks sharper and clearer.

\subsection{Cross-domain evaluation}


\begin{table}[tbhp]
\centering
\caption{Quantitative results for cross-domain evaluation.} \vspace{0.2cm}
\setlength\tabcolsep{3pt}
\begin{tabular}{|c|c|c|c|c|c|c|c|c|c|}
\hline
& \multicolumn{2}{c|}{D-Hazy-NYU~\cite{ancuti2016d}} & \multicolumn{2}{c|}{D-Hazy-MB~\cite{ancuti2016d}} & \multicolumn{2}{c|}{I-Haze~\cite{ancuti2018ihaze}} & \multicolumn{2}{c|}{O-Haze~\cite{ancuti2018ohaze}} \\
\hline
& PSNR & SSIM & PSNR & SSIM & PSNR & SSIM & PSNR & SSIM\\
\hline
DCP~\cite{he2011single} & 11.56 & 0.6695 & 12.13 & 0.6752 & 13.41 & 0.4930 & 17.01& 0.4875\\
\hline
CAP~\cite{zhu2015fast} & 13.29 & 0.7266 & \underline{14.36} & \textbf{0.7526} & 15.27 & 0.5603 & 16.68 & 0.4810 \\
\hline
DehazeNet~\cite{cai2016dehazenet} & 13.02 & 0.7256 & 13.78 & 0.7342 & \textbf{16.73} & \underline{0.6263} & \textbf{17.90} & \textbf{0.5514} \\
\hline
 MSCNN~\cite{ren2016single} & \underline{13.67} & \underline{0.7413} & 13.97 & \underline{0.7488} & 15.93 & 0.5896 & 16.27 & 0.4947 \\
 \hline
AOD-Net~\cite{li2017aod} & 12.44 & 0.7147 & 13.48 & 0.7470 & 15.00 & 0.5828 & 16.22 & 0.4142\\
\hline
 Ours & \textbf{18.11} & \textbf{0.8268} & \textbf{15.63} & 0.7338 & \underline{16.04} & \textbf{0.6332}  & \underline{17.46} & \underline{0.5337}
\\
\hline
\end{tabular}%
\label{tab:cross}%
\vspace{-0.3cm}
\end{table}%

In this section, we focus on the cross-domain performance by evaluating our network trained on RESIDE-standard \cite{li2017reside} on the cross domain datasets, D-Hazy~\cite{ancuti2016d}, I-Haze~\cite{ancuti2018ihaze}  and O-Haze~\cite{ancuti2018ohaze}. We compare with baseline methods that have publicly available code, and these are strong baselines according to benchmark evaluation in \cref{tab:soa}. For learning-based methods DehazeNet~\cite{cai2016dehazenet},  MSCNN~\cite{ren2016single}, and AOD-Net~\cite{li2017aod}, we use the best model the authors have released. MSCNN~\cite{ren2016single} and AOD-Net~\cite{li2017aod} are trained with synthetic images similar to RESIDE-standard, while DehazeNet~\cite{cai2016dehazenet} is trained with patches of web images. 

We present the quantitative results in \cref{tab:cross}, where we use bold to label the best results and underline to label the second best results. Our approach achieves best results, or close to the best results for all the cross-domain evaluations. Our first observation is that the learning-based methods~\cite{cai2016dehazenet,ren2016single,li2017aod}, including ours, generalize reasonably well and perform equally or better than the prior-based methods~\cite{he2011single,zhu2015fast}. 

Our network performs well on the cross-domain D-Hazy dataset~\cite{ancuti2016d}. 
Particularly, our approach outperforms all baseline methods by a large margin on the images synthesized from NYU depth dataset. D-Hazy dataset is synthesized by the same clear images as our training data RESIDE-standard, but uses different parameters of the atmospheric scattering model. Our trained network has effectively captured the statistics of the deep features of the desired clear images. 

I-Haze~\cite{ancuti2018ihaze}  and O-Haze~\cite{ancuti2018ohaze} images look quite different from our training images, and our network may have difficulty to infer the exact statistics of deep features for these images. DehazeNet~\cite{cai2016dehazenet} may have gained some advantage on these two datasets because it is trained on patches of web images. Our approach still produces competitive results compared with DehazeNet~\cite{cai2016dehazenet}, and outperforms all the other baselines. Notice again that our network does not use the powerful  atmospheric scattering model, and is only trained on a limited number of indoor synthetic images. The cross-domain evaluation further demonstrates the power of \emph{deep} features and instance normalization in our approach.

\vspace{-0.1cm}
\subsection{Qualitative evaluation}

\begin{figure}[tbhp]
\vspace{0.1cm}
\centerline{
\includegraphics[width=1\linewidth]{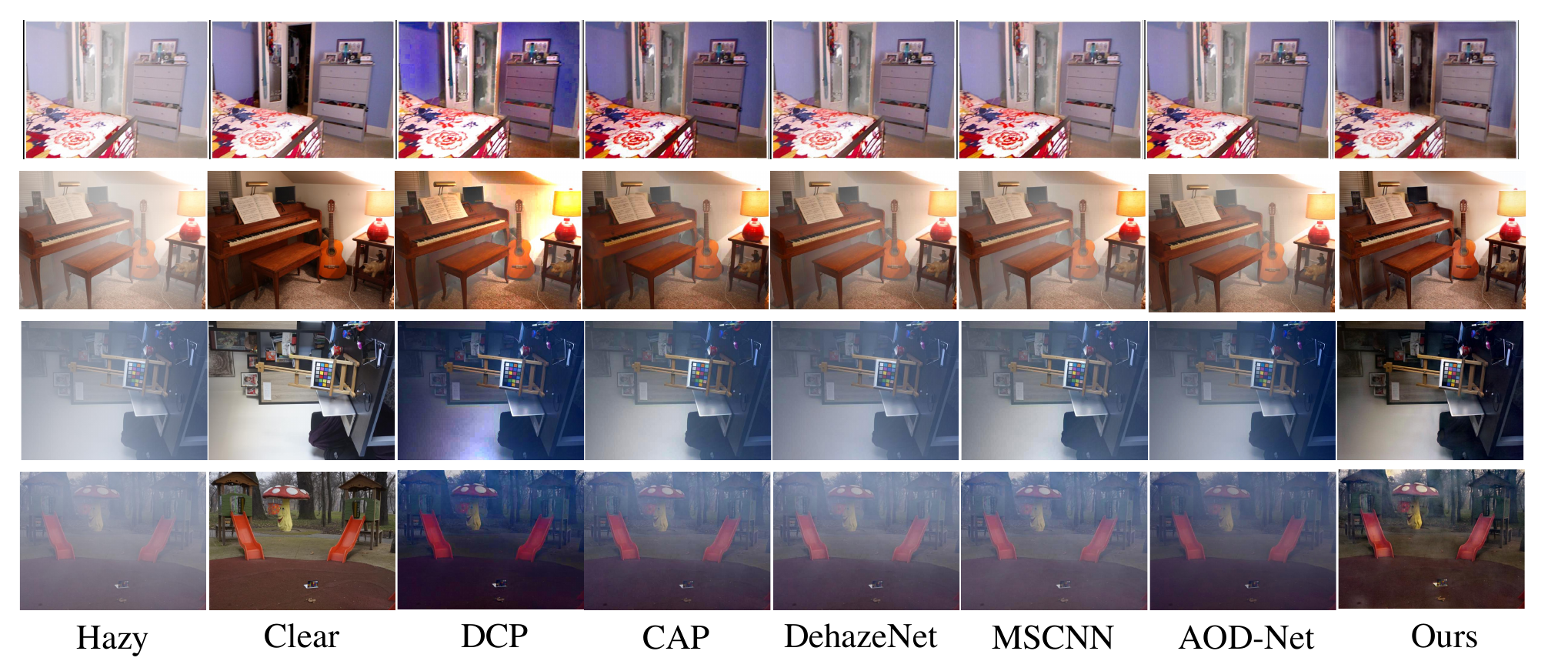}
}
\vspace{0.25cm}
\caption{
Qualitative evaluation on cross-domain dataset. The four examples are from D-Hazy-NYU~\cite{ancuti2016d}, D-Hazy-MB~\cite{ancuti2016d}, I-Haze~\cite{ancuti2018ihaze}  and O-Haze~\cite{ancuti2018ohaze}, respectively. Best viewed in color and zoomed in. }
\label{fig:vis}
\end{figure}

We present  qualitative results from cross-domain evaluation in \cref{fig:vis}. The images are  from D-Hazy-NYU~\cite{ancuti2016d}, D-Hazy-MB~\cite{ancuti2016d}, I-Haze~\cite{ancuti2018ihaze}  and O-Haze~\cite{ancuti2018ohaze}, respectively. We show the hazy image and groundtruth clear image, and compare our results with DCP~\cite{he2011single},  CAP~\cite{zhu2015fast}, DehazeNet~\cite{cai2016dehazenet},  MSCNN~\cite{ren2016single}, and AOD-Net~\cite{li2017aod}. We use the best released model for the learning-based baselines~\cite{cai2016dehazenet,ren2016single,li2017aod}, and train our network on RESIDE-standard~\cite{li2017reside}.

Our network makes the best efforts to remove haze and recover the real color of images, as shown in \cref{fig:vis}. The results of baselines still have a large amount of undesired haze and look blurry (row 2,3,4). Particularly, the baselines have difficulty in dark areas of the image, and DCP also has difficulty in area of white and blue walls (row 1,3). For the outdoor image (row 4), our network produces a little artifact due to the significant domain difference between the desired images and the training indoor images. Use regularizers such as total variation \cite{rudin1992nonlinear} may help reduce these artifacts, and we plan to investigate it in the future. Our simple yet effective network has generated visually appealing results, without depending on extra constraints like the atmospheric scattering model.

\vspace{0.2cm}
\section{Discussion}
We proposed a simple yet effective end-to-end system for single image dehazing.
 Our network has an encoder-decoder architecutre with skip connections. 
 We manipulated the statistics of deep features extracted by pre-trained VGG net and demonstrated the effectiveness of instance normalization for image dehazing. 
 Moreover,  without explicitly using the atmospheric scattering model, our approach outperforms previous methods by a large margin on the benchmark datasets. 
  Notice that both the training and testing data are generated by the atmospheric scattering model, and the baseline methods all explicitly use the model. 
  Our network effectively learns the transformation from hazy image to clear image with limited synthetic data, and generalizes reasonably well. 
  
The atmospheric scattering model is powerful and has been successfully deployed for image dehazing in the past decade. 
However, the atmospheric scattering model, as a simplified model, also constrained the learnable components to be ``linearly'' combined by element-wise multiplication and summation, which may not be ideal for training deep models. 
Our study sheds light on the power of \emph{deep} neural networks and the \emph{deep} features extracted by pre-trained network for single image dehazing, and encourages the rethinking on how to effectively exploit the physical model for haze.
How will physical model help when training powerful deep networks? 
It is still an open question, and our approach serves as a strong baseline for future study.  

Our network outperforms state-of-the-art methods by a large margin on the benchmark dataset, and achieves competitive results on cross-domain evaluation. The key idea of our approach is to apply instance normalization to shift the statistics of deep features for image dehazing. For cross-domain evaluation, it may be difficult to effectively infer the desired statistics of deep features of clear images that is quite different from the training data. Our generalization ability can be significantly improved by training from large-scale natural images. In the future, we will explore adversarial training to use unpaired hazy and clear images that are easier to collect from the web.

\bibliographystyle{plainnat}
\small
\bibliography{dehaze,style}
\end{document}